\documentclass[10pt,twocolumn,letterpaper]{article}
\usepackage{wacv}              

\usepackage{multirow}
\usepackage{graphicx}
\usepackage{pgffor}

\definecolor{wacvblue}{rgb}{0.21,0.49,0.74}
\usepackage[pagebackref,breaklinks,colorlinks,allcolors=wacvblue]{hyperref}

\def\wacvPaperID{3087} 
\def\confName{WACV}
\def\confYear{2026}

\usepackage[accsupp]{axessibility}
\title{
Learning Compact Video Representations for Efficient Long-form Video Understanding in Large Multimodal Models}

\author{Yuxiao Chen, Jue Wang, Zhikang Zhang, Jingru Yi, Xu Zhang, Yang Zou\\ Zhaowei Cai, Jianbo Yuan, Xinyu Li, Hao Yang, Davide Modolo \\ 
Amazon AGI
}

\begin{document}
\maketitle
\begin{abstract}
With recent advancements in video backbone architectures, combined with the remarkable achievements of large language models (LLMs), the analysis of long-form videos spanning tens of minutes has become both feasible and increasingly prevalent. However, the inherently redundant nature of video sequences poses significant challenges for contemporary state-of-the-art models. These challenges stem from two primary aspects: 1) efficiently incorporating a larger number of frames within memory constraints, and 2) extracting discriminative information from the vast volume of input data. In this paper, we introduce a novel end-to-end schema for long-form video understanding, which includes an information-density-based adaptive video sampler (AVS) and an autoencoder-based spatiotemporal video compressor (SVC) integrated with a multimodal large language model (MLLM). Our proposed system offers two major advantages: it adaptively and effectively captures essential information from video sequences of varying durations, and it achieves high compression rates while preserving crucial discriminative information. The proposed framework demonstrates promising performance across various benchmarks, excelling in both long-form video understanding tasks and standard video understanding benchmarks. These results underscore the versatility and efficacy of our approach, particularly in managing the complexities of prolonged video sequences.
\end{abstract}
    
\section{Introduction}
\label{sec:intro}
\begin{quote}
\textit{"The whole is greater than the sum of its parts."} 
\vspace{-0.3cm}
\begin{flushright}
--- Aristotle
\end{flushright}
\vspace{-0.5cm}
\noindent\rule{0.4\textwidth}{0.4pt}
\end{quote}

\begin{figure}[thb]
\centering
\includegraphics[width=\linewidth]{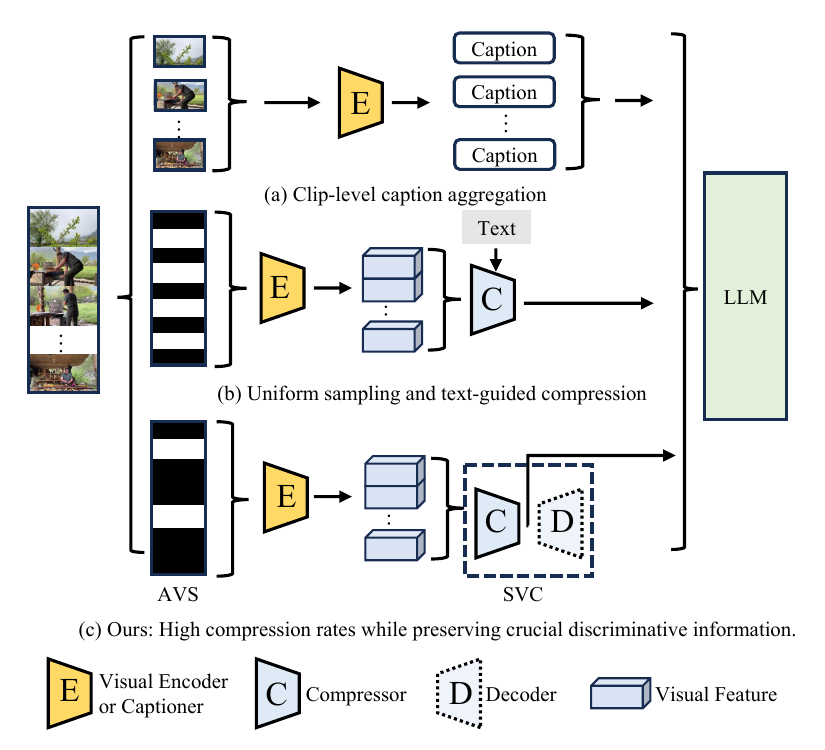}
  \caption{An overview of our method and previous works, showing the different way of modeling long-form video with LLM. Specifically, (a) interpret clip into clip-level caption and aggregate via LLM in the linguistic space, (b) uniformly sample the video frame and leverage paired text to compress video tokens, and (c) our proposed method that leverage adaptive video sampler (AVS) and autoencoder based spatiotemporal video compressor (SVC).}
\label{fig:fig_1}
\vspace{-5pt}
\end{figure}

\noindent Long-form video understanding has emerged as a hot topic in the research community, presenting significant challenges due to its complexity and dynamic nature, and remains an unsolved problem despite ongoing advancements. Conventional approaches~\cite{wu2021lvu,tang2019coin,kuehne2014breakfast,wang2023s5, islam2022vis4mer} often focus on classification tasks, where the goal is to map an entire video to one or a few predefined ground truth labels. These methods typically treat the video as a single, unified entity, aiming to assign a label that best represents the overall content. Although effective in some contexts~\cite{wu2021lvu,tang2019coin,kuehne2014breakfast}, these approaches are insufficient to understand long-form video, as they do not capture the complexity and contextual relationships unfolded across multiple scenes over time.

Recently, multimodal large language models (MLLMs)~\cite{shen2024longvu, lin2023video, maaz2023video, xu2024slowfast, lin2023video, li2024llava, song2024moviechat} have made a significant impact on the field of video understanding, showcasing remarkable emergent capabilities across a variety of video-related tasks, such as perception, common sense reasoning, and open dialogue~\cite{li2024mvbench, mangalam2023egoschema, patraucean2024perception, yu2019activitynet,xiao2021next}. However, many of these models face challenges in effectively learning and modeling long-form video. This difficulty arises from the redundant nature of video sequences, which often results in an overwhelming number of visual tokens, leading to substantial computational overhead in LLMs. RoPe~\cite{su2024roformer} provides a solution to handle the extended context during inference. However, the complexity of a long-form video sequence requires a more effective encoding solution during training.

To address these challenges, one successful attempt~\cite{islam2024video, wang2024videoagent, zhang2023simple,liao2024videoinsta}, illustrated in Figure~\ref{fig:fig_1} (a), is to represent the long-video hierarchy with the natural language, which segments the long video sequences into several short clips and produces video segment captions. These short captions are subsequently fed into the LLMs for inference. However, these models interpret video clips into natural languages in the early stage, leading to the loss of much low-level visual information. Additionally, aggregated hallucinations from segment-level captions may result in poor generalizability and performance across various tasks and models. Another direction~\cite{song2024moviechat,li2025llama} focuses on reducing the large number of tokens in long-form videos by introducing a learnable compression or token merging module, or a memory bank linked to the LLM, as shown in Figure~\ref{fig:fig_1} (b). However, scaling these methods is challenging, as they require a large volume of video-text pairs, with the linguistic information serving as guidance for training the compressor. Finally, parameter-free pooling operations (such as average pooling) have been proven effective~\cite{shen2024longvu, lin2023video, maaz2023video, xu2024slowfast, lin2023video, li2024llava, song2024moviechat}, but information loss remains unavoidable and is even more severe in long-form videos, where the diversity within video sequences is greater than in short videos.

In this paper, we introduce a comprehensive schema for long-form video understanding, covering the entire pipeline, from sampling to compression to high-level interpretation. This system includes: 1) an Adaptive Video Sampler (AVS) that selects video frames based on information density, 2) an Auto-Encoder based~\cite{rumelhart1986learning} Spatiotemporal Video Compressor (SVC) that can be effectively trained with video-only data, and 3) a MLLM that integrates seamlessly with both the compressor and sampler. The AVS works jointly with SVC delivering 64-time compression to save the token budget for MLLMs. An overview of our proposed system is illustrated in Figure~\ref{fig:fig_1} (c). Our approach significantly reduces the token budget by 64 times, enabling MLLMs to process hours-long video sequences in their entirety. The proposed AVS and SVC modules ensure a sufficient capture of discriminative information while considerably expanding the perception field of the MLLM in long-form videos.

We conducted extensive experiments in a diverse range of tasks and problem settings to evaluate the proposed method. Our approach significantly outperforms state-of-the-art methods on various video benchmark downstream tasks while demonstrating superior efficiency by using substantially fewer visual tokens and reducing computational overhead. Notably, our method surpasses Llava-OV~\cite{li2024llava} by 2.6\% and 3.3\% on the EgoSchema and PercepTest tasks, respectively, while utilizing 80\% fewer visual tokens. Our contributions are as follows.
\begin{itemize}[align=parleft,left=10pt..1em, leftmargin=*]
\itemsep0em
    \item We proposed a novel long-form video understanding schema to equip MLLM, which includes two novel components: Adaptive Video Sampler (AVS) which samples frames based on the information density, and Spatio-temporal Video Compressor (SVC) that is an AE-based video compressor. 
    \item The proposed AVS and SVC work jointly to deliver 64-time compression ratio, which significantly save the token budget in MLLM while preserving the discriminative information of the long-form videos.
    \item Compared to baseline and competitive counterparts, we demonstrate the effectiveness of the proposed elements and achieve performance on par with the SoTA methods in various video understanding benchmarks, while only utilizing 20\% video tokens compared to the previous SoTA method~\cite{li2024llava}.
\end{itemize}


\section{Related Work}

\subsection{MLLMs in Long-form videos}
In contrast to conventional long-form video understanding tasks that classify entire long video as one or a few ground truth labels~\cite{wu2021lvu,tang2019coin,kuehne2014breakfast,wang2023s5, islam2022vis4mer, wang2025gexia, wang2019discriminative, lee2024video, hori2018multimodal, chen2024learning, bao2025exploiting}, Multi-modal Large Language Models (MLLMs) interpret long-form video within a well-defined linguistic space, thus expanding the scope of video understanding to include tasks such as contextual reasoning, open-ended dialogue about video content, and complex question-answering across long-video sequences. 
However, the high computational cost of transformer-based LLMs poses a significant barrier to effectively processing entire sequences of long-form videos within memory constraints. To address this issue, one solution is to divide the entire long-form video into short clips and generate clip-level captions, which will be hierarchically aggregated as long-form video representation via LLM~\cite{islam2024video, wang2024videoagent, zhang2023simple,liao2024videoinsta}. However, these models often lose significant low-level visual information, and the accumulated hallucinations from segment-level captions can lead to poor generalizability and reduced performance across various tasks and models. To this end, an efficient and effective video encoding scheme is required in long-form video understanding tasks. 

\subsection{Long-form Video Encoding}
\noindent\textbf{Token Reduction} Token reduction is one of the most effective methods to relieve computational burden. It can be further categorized as \textit{1) Token Selection}, \textit{2) Token Merging} and \textit{3) Token Compression}. 

\noindent\textit{Token Selection}: Token selection is a common strategy for improving model efficiency, utilizing lightweight modules to retain only the most "useful" tokens while discarding those considered less essential. Typical works, such as STTS~\citep{wang2021efficient}, AdaViT~\citep{meng2022adavit} and similar methods~\citep{wang2021efficient, meng2022adavit, rao2021dynamicvit, liang2022not} introduce a selection mechanism to assign importance scores to each token and selects the top-K important ones. However, it is impractical for a lightweight selection module to select discriminative information from a large volume of long-form video data without risking a loss of contextual information.

\noindent\textit{Token Merging}: Originally proposed by~\citep{bolya2022tome}, token merging is designed to enhance throughput in vision transformers without the need for further training. The merging technique splits visual tokens into two equal groups; for each token on the edge of one group, it identifies the closest matching token in the other group, merges them using average pooling, and then recombines the groups. Although this method is effective and straightforward, its use has been mainly limited to the image and short video domains~\citep{ren2023testa, bolya2023tomesd, li2023vidtome}. This is due to the high diversity in long-form videos, which often leads to mismatches in a similarity-based merging system. More practically, a sliding window-based averaging pooling cubic is widely adapted in recent long-form video understanding works~\cite{maaz2023video, xu2024slowfast,lin2023video,li2024llava,cheng2024videollama}. In particular, Video-ChatGPT~\cite{lin2023video} and SlowFast-LLaVA~\cite{xu2024slowfast} employ two distinct pooling strategies across spatial and temporal dimensions to capture object and motion information, respectively. However, these pooling strategies treat each video frame equally, which may result in information distortion when two dissimilar candidates are combined. Effectively reducing redundant visual tokens while preserving key information remains a challenge and further exploration is needed. 

\noindent\textit{Token Compression}: Recent works~\cite{wiles2022compressed, li2025llama, song2024moviechat,shen2024longvu,yu2023language} have successfully leveraged a compressor to reduce the number of visual tokens while preserving useful information in various applications, such as classification, generation, and general understanding with LLMs. For example, Wiles et al.~\cite{wiles2022compressed} leverage the VQ-VAE encoder-decoder model~\cite{van2017neural} to apply augmentations
directly on the latent codes of compressed video and use them for classification tasks. Yu et al. propose MAGVIT-v2~\cite{yu2023language}, which generates concise and expressive tokens for both videos and images utilizing a shared VQ-VAE codebook. In the field of long-form video understanding, MovieChat~\cite{song2024moviechat} introduces a recurrent memory bank to consolidate excessive video tokens, while LLaMA-VID~\cite{li2025llama} uses cross attention to represent each video frame as two tokens. Furthermore, LongVU~\cite{shen2024longvu} utilizes DINOv2~\cite{oquab2023dinov2} features to eliminate redundant frames that exhibit high similarity. These works either rely on external prior or well-structured video and text pairs, which may easily introduce bias or prevent the algorithm scaling up. In this work, we argue that the token reduction mechanism should be implemented as a holistic pipeline that jointly considers video sampling and compression problems as a whole. To this end, we propose a simple yet effective adaptive video sampler as well as a video compressor that delivers a 64-time compression ratio over long-form video while preserving discriminative information.

\noindent\textbf{Video Backbone} Recently, Mamba~\cite{gu2023mamba} and its huge family~\cite{wang2023s5,liu2024vision,yang2024vivim,chen2024video} received promising results in various long context modeling tasks. Specifically, it utilizes a state-space sequence model with a structured convolutional kernel to efficiently capture long-context information, achieving linear complexity. However, stringent initialization conditions make it difficult for Mamba to be trained in large scale with MLLMs~\cite{liu2024vision}. As a result, we follow previous works~\cite{shen2024longvu, lin2023video, maaz2023video, xu2024slowfast, lin2023video, li2024llava, song2024moviechat} by using ViT~\cite{dosovitskiy2020image} as the video backbone. The exploration of Mamba with MLLMs will be left for future work.
\section{Methodology}

\begin{figure*}[t]
\centering
\includegraphics[width=\linewidth]{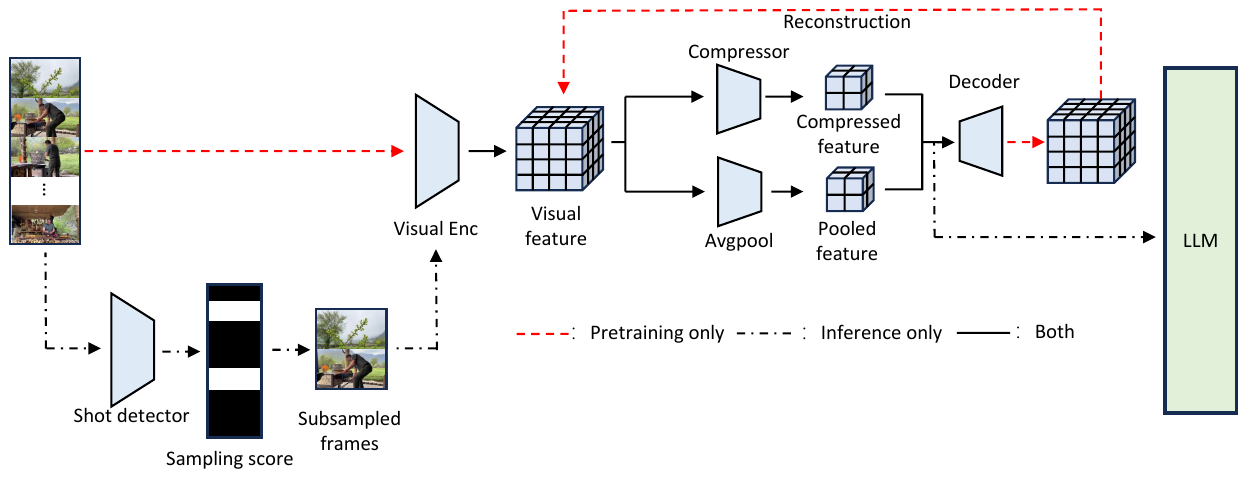}
  \caption{Overview of the proposed method.}
\label{fig:fig_2}
\vspace{-5pt}
\end{figure*}

\subsection{Preliminaries}
\noindent\textbf{MLLM For Long-form Video Understanding}\\
\noindent Multimodal Large Language Models (MLLMs) process vision data using LLMs to interpret visual information extracted from inputs alongside task-specific text prompts (e.g., visual questions about visual content), ultimately generating the corresponding responses~\cite{cheng2024videollama,li2025llama,liu2024visual}.

To be specific, given an input video $V$ as an example,  a set of frames, denoted as  $X$, are first \textit{uniformly} sampled form it. A pretrained visual encoder $\mathcal{E}$ then extracts visual token embeddings, denoted as $f$, from these sampled frames:

\begin{equation}
f = \mathcal{E}(X), \quad f \in \mathcal{R}^{T \times H \times W \times C}
\end{equation}
where $T$ represents the frame number,  $H$, and $W$ are height and width of the extracted feature tensor, respectively.

The LLM subsequently processes the visual tokens with the text question ($Q$) to generate the answer $A$.

\begin{equation}
A = LLM(P(f), Q)
\label{eq:no_compress}
\end{equation}
where $P$ is a projector, such as the MLP module, that maps visual token features into the input space of the LLM.

However, in the context of long-form video, the volume of $f$ increases significantly, which impacts the capacity of MLLM~\footnote{The complexity of transformer-based LLMs is quadratic in relation to input length.} in terms of both efficiency and effectiveness. Additionally, uniform sampling combined with standard average pooling is suboptimal for long-form video, as redundant and irrelevant frames may consume the limited token budget of LLMs, ultimately reducing their ability to capture complex long-term dependencies. Motivated by these, we propose a long-form video understanding framework in this work, offering solutions and best practices for the research community.

\subsection{Proposed Framework}
In this paper, we propose a novel long-form video understanding framework, addressing the aforementioned challenges, from frame loading strategy to video token compression technique. Our framework consists of an Adaptive Video Sampler (AVS), a ViT-based video encoder, an autoencoder-based Spatiotemporal Video Compressor (SVC) and the recent popular LLM, QWen2~\cite{yang2024qwen2}. An illustration is provided in Figure~\ref{fig:fig_2}.

\subsubsection{Adaptive Video Sampling}
\label{sec:afs}

In long-form videos, the discriminative information is not evenly distributed. Given the tight token budget of LLMs, it is important to carefully pick visual tokens from the entire sequence. To this end, we propose an information-density-based frame sampler that can adaptively select informative video frames from a long video sequence. We follow the decomposition rule of a movie, borrowing the concept of chapter, scene, and shot in the movie. It is assumed that each long-video sequence consists of serval continues information tubelets, with intra-tubelet information being generally homogeneous, while inter-tubelet data distributions vary significantly. As a result, the gradient of information shifts is crucial for identifying where discriminative information lies in the long-form video. In this work, we employ a shot boundary detection module as a core component of our adaptive frame sampler to capture dynamic moments that contain key information.
 
Specifically, we first feed the entire video $V$ to a shot boundary detector $S$, which outputs a confidence score that indicates the likelihood of a content change with respect to each frame:

\begin{equation}
      S = \mathcal{S_D}(V), \quad S = \{s_1, s_2, ..., s_N\}
\end{equation}
where $s_i$ is the confidence score for the $i$-th frame, and $N$ is the number of frames.

We then apply the non-maximum suppression (NMS) scheme to filter out redundant detections:

\begin{equation} 
S' = \mathrm{NMS}(S) 
\end{equation}

Next, we sample frames with the top-k highest confidence scores and sort them temporally for video understanding:
\begin{equation} K = \operatorname{Sort}(\operatorname{Top}_k(S')) \end{equation} where $K$ represent the index of sampled frames

\subsubsection{Autoencoder-based Video Compressor}
\label{sec:ae}
With the smart sampling strategy, the perception field of LLMs is still constrained by the massive number of visual tokens per frame. In the area of image and short-form videos, average pooling is widely used as an effective way to reduce redundant tokens~\cite{maaz2023video, xu2024slowfast,lin2023video,li2024llava,cheng2024videollama}. However, it is argued that the average pooling is not ideal in long-form video understanding, as the high diversity among sparsely sampled frames can lead to information distortion when forced into aggregation. Unlike previous works that rely on knowledge priors or well-aligned video-text pairs to compress video tokens, we propose an autoencoder-based video compressor in this work. This compressor condenses raw video feature representations into a compact latent space, reducing the number of visual tokens while retaining essential information. The autoencoder architecture ensures the scalability of the compressor, allowing it to be trained using video data alone. In~\cite{cheng2024videollama}, it is pointed out that the average pooling demonstrates superior performance than the convolution-based compressor. In this work, we show that the convolution-based compressor with autoencoder pretraining significantly outperform the average pooling.

To be specific, the compressor model $\mathcal{C}$ removes redundant information among tokens by compressing raw video features $f$ to a compact latent space, 

\begin{equation}
h = \mathcal{C}(f), \quad h \in \mathcal{R}^{t \times h \times w \times c}
\end{equation}
where $t$, $h$, $w$ are the temporal length, height, and width of the feature tensor in the latent space, $t \leq T$, $h \leq H$, and $w \leq W$.

The decoder $\mathcal{D}$ model then reconstructs the raw feature representation from $h$:
\begin{equation}
    \hat{f} = \mathcal{D}(h), \quad  \hat{f} \in \mathcal{R}^{T \times H \times W \times C}
\end{equation}

The compressor $\mathcal{C}$ and the video decoder $\mathcal{D}$ are trained together to minimize the difference between $\hat{f}$ and $f$ using the mean absolute loss:

\begin{equation}
\mathcal{L}_{rec} = |f - \hat{f}| \
= |\mathcal{E}(X) -  \mathcal{D}(\mathcal{C}({\mathcal{E}}(X))) |
\label{eq:ae}
\end{equation}

Note that when optimizing $\mathcal{C}$ and $\mathcal{D}$ to minimize $\mathcal{L}_{rec}$, the visual encoder $\mathcal{E}$ remains fixed. The loss forces the encoder to retrain all necessary information in the compressed representation during the compression process. 

\textit{The reconstruction loss also enables us to pre-train the compressor $\mathcal{C}$ using only video data, thereby learning valuable prior knowledge. In contrast, previous methods train the compressor alongside LLMs for next-text-token prediction tasks, which require costly visual-text pairs for training. The limited scale of such data may constrain the compressor's performance.}

After pretraining, we leverage the compressor $\mathcal{C}$ with the visual encoder $\mathcal{E}$ to extract compressed video representation from the video, which is then combined with a text question $Q$ and fed into the LLM for response inference.

\begin{equation}
A = LLM(P(\mathcal{C}(\mathcal{E}(X)), Q)
\end{equation}

\noindent\textbf{Residual Latent Space Constraint:} We empirically observe that aligning the compressor pretrained using Equation~\ref{eq:ae} with LLMs presents significant challenges. This difficulty stems from the lack of constraints within the compact latent space. Consequently, the learned compressor has limited generalization capabilities and may encode unseen data into gaps or "holes" within the latent space, leading to a loss of meaningful representation. To mitigate this issue, we propose incorporating the average pooled feature of $X$ with $h$ as constraints, redefining the latent feature $h$ and reconstruction loss $\mathcal{L}_{rec}$ as follows:

\begin{equation}
    h = \mathcal{C}(f) + avgpool_{3D}(X)
\end{equation}
\begin{equation}
    \mathcal{L}_{rec} = |\mathcal{E}(X) -  \mathcal{D}(\mathcal{C}({\mathcal{E}}(X)) + Avgpool_{3D}(X)) |
\end{equation}

With the proposed constraint, the compressor focuses on learning the residual (lost information) from the average pooling process, thus reducing the learning complexity of the autoencoder while implicitly ensuring that the learned feature $\mathcal{C}(f)$ aligns with the same space as $avgpool_{3D}(X)$. Additionally, compared to Variational Autoencoders (VAEs), which enforce the latent feature space to follow a Gaussian distribution, our constraint achieves superior performance by eliminating nondeterministic behavior in the latent space, which can increase the challenges of learning for the autoencoder (see experiments for more results).

\noindent \textbf{Compression Ratio:} Compared with the setting where the output features from the visual encoder are used directly as input to the LLM (see equation~\ref{eq:no_compress}), the compressor reduces the token number by factors of $\frac{T}{t}$, $\frac{H}{h}$, and $\frac{W}{w}$ for the temporal, height, and width dimensions, respectively. This results in a total compression ratio of $\frac{T}{t} \times\frac{H}{h}\times\frac{W}{w}$ times.

\noindent\textbf{Light-weight 3D convolutional kernel:}
We implement $C$ using cascaded convolutional residual blocks. \textit{We chose a convolution-based compressor for its simplicity and superior computational efficiency compared to transformer-based models such as Perceiver. Additionally, convolution operations effectively utilize the inductive bias that assumes that local visual tokens have high redundancy}. To reduce the parameter size, we decompose the 3D convolution operation into a 2D spatial convolution and a 1D temporal convolution. Additionally, we apply a channel-wise bottleneck mechanism when the 2D spatial convolution maps the feature to a lower dimension than the original, while the temporal convolution restores the feature to its original dimension. The downsampling is achieved through the convolutional strides $( s_h, s_w)$ for the 2D convolution and $s_t$ for the 1D convolution.

\subsection{Implementation Details}
We adopt a four-stage training recipe. (1) In the first stage, we train a VAE-based video compressor with supervision of construction loss. (2) Next, a projector between the video features and LLM is trained with our filtered ShutterStock data ($\sim$3M) to align the visual features from visual space to language space. (3) In the third step, we unfreeze the LLM weights and further jointly train the LLM and projector with $\sim$8M data which is collected from subset of ShutterStock, Ego4D \cite{grauman2022ego4d}, Breakfast \cite{kuehne2014language}, AVA \cite{li2020ava}, Vatex \cite{wang2019vatex}, Something-somethingv2 \cite{goyal2017something}, Kinetics \cite{carreira2018short} datasets. (4) In the last step, we finetune the model with our supervised finetuning (SFT) dataset which is collected from NextQA \cite{xiao2021next}, CLEVRER \cite{yi2019clevrer}, PerceptionTest \cite{patraucean2024perception}, and Egoschema \cite{mangalam2023egoschema}. The SFT data adds up to 1M scale.  
\section{Experimental Results}
\begin{table*}[t]
\footnotesize
\centering
\begin{tabular}{lccccccc}
\toprule
Model       & \multicolumn{2}{c}{EgoSchema}   & NextQA   & ActivityNetQA & MLVU & MVbench& PerceptionTest\\ 
                       & fullset            & subset          & mc     & acc                &  avg & test  &val\\ 
\midrule
VideoChat~\cite{li2023videochat}              & -              & -             & -      & 26.5                   & - & - & - \\
LLaMA Adapter~\cite{li2023videochat}          & -              & -             & -      & 34.2                  & - & - & - \\
Video-ChatGPT~\cite{li2023videochat} (ACL’24)  & -              & -             & -      & 35.2                  & - & - & -  \\
LLaMA-VID~\cite{li2023llamavidimageworth2} (ECCV’24)     & -              & -             & -      & 47.5                  & - & - & -  \\
Movie\_Chat~\cite{song2024moviechat} (CVPR’24)   & -              & -             & -      & 45.7                  & - & - & -  \\
VideoAgent~\cite{wang2024videoagent} (ECCV’24)    & 54.1           & 60.2          & 71.3   & -                      & - & - & -  \\
LLoVi~\cite{zhang2023llovi} (EMNLP’24)        & 52.2           & 58.8          & 73.8   & -                      & - & - & -  \\
VideoINSTA~\cite{liao2024videoinsta} (EMNLP’24)   & -              & \underline{65.0} & 72.3   & -             & -             & - & -  \\ 
\midrule
ShareGPT4V~\cite{chen2024sharegpt4video} & - & - & - & - & 46.4 & 51.2 & -
\\
LLaVA-NeXT-Video-7B~\cite{zhang2024llavanextvideo} & 43.9 & - & - & - & - & 33.7 & - 
\\
LongVA-7B~\cite{zhang2024long} & -   & - & 68.3  & 50.0           & 56.3      & -            & -             \\
LLava-OneVision-7B~\cite{li2024llava}   & \underline{60.1} &- & \textbf{79.4} & \textbf{56.6}        & \textbf{64.7}      & \underline{56.7}           & \underline{57.1}          \\
\textbf{Ours}     & \textbf{62.7}  & \textbf{69.6}  & \underline{78.7}    & \underline{52.7}       & \underline{59.6}      & \textbf{58.8}            & \textbf{60.4} \\
\bottomrule
\end{tabular}
\caption{Comparison with state-of-the-art MLLM-based methods on both long-form video understanding and general video understanding tasks. The bold and underline texts represent the best and the second best performances, respectively.}
\label{tbl:sota}
\end{table*}

\subsection{Benchmark and Metric}
We conduct experiments on general general video understanding benchmarks covering a wide range of video lengths and video types, including: 
\textbf{PerceptionTest~\cite{patraucean2024perception}} a multimodal benchmark designed to show perceptually interesting situations and defines multiple tasks. Following~\cite{patraucean2024perception} we report the accuracy of multichoice questions (MCQ).
\textbf{ActivityNet-QA}~\cite{yu2019activitynet} contains 58,000 human-annotated QA pairs of 5,800 videos derived from the popular ActivityNet dataset. The dataset provides a benchmark for testing the performance of VideoQA models on spatio-temporal reasoning. Following \cite{wang2024videoagent}, we report both accuracy and LLM based answer matching scores for above datasets. 
The \textbf{MVBench~\cite{li2024mvbench}} is a comprehensive Multi-modal Video understanding benchmark with videos ranging from 5s to 35s. MVBench covers 20 challenging video tasks that require a broad spectrum of temporal skills, ranging from perception to cognition. We return the average score on the downstream tracks as suggested by~\cite{li2024mvbench}. The \textbf{NExTQA}~\cite{xiao2021next} is a VideoQA benchmark that focuses on the reasoning of causal and temporal actions reasoning and object interactions understanding in daily activities. It contains 48K multi-choice questions with average duration of 44s. For these datasets, we report the multi-choice accuracy following previous practice~\cite{wang2024videoagent}. 
For long video understanding, we choose \textbf{EgoSchema}~\cite{mangalam2023egoschema} which is derived from Ego4D \cite{grauman2022ego4d}, consists of over 5000 human-curated multiple choice question answer pairs, with videos of three-minute duration. And \textbf{MLVU}~\cite{zhou2024mlvu} which is designed for very Long Video Understanding (LVU) tasks. It is constructed from a wide variety of long videos, with lengths ranging from 3 minutes to 2 hours. We report the mean accuracy of multi-choice accuracy on all downstream tasks as suggested by benchmark.

\subsection{SoTA Comparions} 
\noindent\textbf{Comparison with SoTA long-form video understanding methods.} First, we benchmark our method against previous long-form video understanding works on the EgoSchema, NextQA, and ActivityNetQA datasets, with results presented in the upper part of the Table~\ref{tbl:sota}. Our method demonstrates significant performance improvements on all three benchmarks. Notably, our method outperforms VideoAgent~\cite{wang2024videoagent}, LLoVi~\cite{zhang2023llovi}, and VideoINSTA~\cite{liao2024videoinsta} by 8.6\% and 4.9\% on the EgoSchema validation set and testing set. These multi-stage methods, which segment videos into short clips, extract clip captions, and aggregate objects through LLM, suffer from error propagation and information loss during multi-stage aggregation. In contrast, our approach learns to extract information end-to-end from videos. The Adaptive Video Sampler (AVS) efficiently filters out redundant temporal information, while our Spatiotemporal Video Compressor with AE maximally preserves of long-term spatiotemporal information, resulting in superior performance. Our method also outperforms LLaMA-VID and Movie-Chat on the ActivityNet-QA dataset by 4.8\%. The two methods rely on conventional, complex pooling-based token compression strategies, such as LLaMA-VID's Q-former-based compression and Movie-Chat's memory-buffer approach for long-term information aggregation. In contrast, our method excels by providing a more lightweight yet effective solution, with less information loss.

\noindent\textbf{Comparison with SoTA MLLMs.} Furthermore, we evaluate our method against state-of-the-art MLLMs across both general and long-form video understanding benchmarks in the lower part of the Table~\ref{tbl:sota}. For fair comparison, we focus on models using 7B LLM backbones.

Although we had access to only a subset of the Supervised Fine-Tuning (SFT) data used in previous works~\cite{li2024llava}, our model still achieves impressive results across all benchmarks.
Specifically, our approach achieves state-of-the-art performance on three out of six benchmarks and comparable results on the remaining ones, demonstrating the effectiveness of our method.
Notably, our model processes EgoSchema and PerceptionTest using an average of just 1,440 visual tokens, substantially fewer than the approximately 6,000 tokens required by LLava-OV~\cite{li2024llava}. This significant reduction in token usage demonstrates how our elegant design, incorporating the proposed sampler and compression module, achieves both computational efficiency and strong performance.

\begin{table*}[]
\footnotesize
\centering
\begin{tabular}{c|c|c|ccccccc}
\toprule
\multirow{2}{*}{Sampler} & NextQA & MVbench & \multicolumn{7}{c}{MLVU (m-avg)}                                   \\ 
                         & MC     & test    & Avg. & Anomaly Det. & Counting & Ego. & Needle & Ordering & PlotQA \\ 
\midrule
Uniform                  & 77.0   & 50.0    & 51.9 & 0.53         & 0.35     & 0.50 & 0.51   & 0.47     & 0.52   \\
Ours (AVS)                      & \textbf{77.1}   &  \textbf{50.2}       & \textbf{52.6} & \textbf{0.54}         & \textbf{0.37}     & \textbf{0.51} & \textbf{0.52}   &\textbf{ 0.49}     &\textbf{ 0.53}  \\
\bottomrule
\end{tabular}
\caption{Experimental results of our method on NextQA, MLVU and MVbench with different video frame sampling strategies.}
\label{tbl:sampler}
\end{table*}

\begin{table}[t]
\footnotesize
\centering
\begin{tabular}{lccccc}
\toprule
\multicolumn{1}{l}{Compressor} & Ratio & EgoS.    & MLVU          & MVbench       &  \\ 
\midrule
\multicolumn{1}{l}{Percevier}  & 64$\times$  & 63.0          & \textbf{53.7} & 49.8          &  \\
\multicolumn{1}{l}{Avgpool-3D} & 2$\times$4$\times$4   & 63.5          & 51.5          & 48.7          &  \\
\multicolumn{1}{l}{Avgpool-3D} & 4$\times$4$\times$4   & 61.4          & 51.3          & 46.8          &  \\ 

\midrule
\multicolumn{1}{l}{Ours (SVC)}     & 4$\times$4$\times$4   & \textbf{63.6} & 51.9          & \textbf{50.0} &  \\
\bottomrule
\end{tabular}
\caption{Comparison between our proposed AE based VAC and different video compressors.}
\label{tbl:arch}
\end{table}

\begin{table}[h]
\footnotesize
\centering
\begin{tabular}{cccccc}
\toprule
Ratio (T$\times$ W $\times$ H) & EgoS.         & MLVU          & MVbench       & NextQA        &  \\ 
\midrule
2$\times$2$\times$2       &     65.1          &           55.0    &        51.3       &             77.1  &  \\
4$\times$4$\times$4       & 63.6 & 51.9 & 50.0 & 77.0 &  \\
1$\times$8$\times$8       & 60.0          & 50.1          & 44.2          & 73.5          &  \\
4$\times$8$\times$8       & 62.0          & 48.7          & 46.6          &             75.5  &  \\
8$\times$4$\times$4       & 62.3          & 48.0          & 43.3          &               73.3& \\
\bottomrule
\end{tabular}
\caption{Experimental results of our method with different video compression ratio.}
\label{tbl:cmp_ratio}
\end{table}

\subsection{Ablation Study}
To validate our design choices and demonstrate the effectiveness of the proposed modules, we conducted ablation studies across four established long-form video benchmarks ~\cite{mangalam2023egoschema, zhou2024mlvu, li2024mvbench, xiao2021next}. 

\begin{figure*}[t]
\centering
\includegraphics[width=\linewidth]{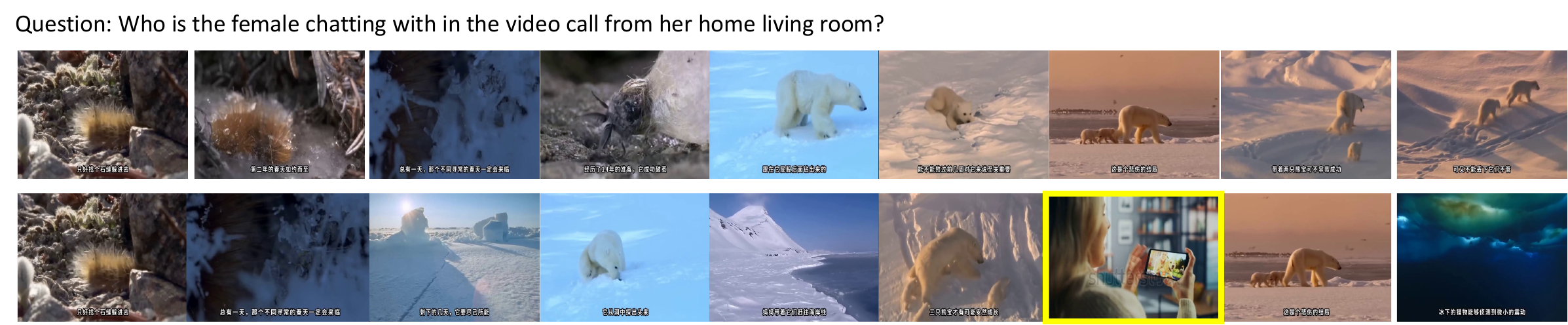}
  \caption{Examples of sampled frames using uniform sampling (top) compared to our AVS (bottom). AVS successfully locate the key frame to answer the question.}
\label{fig:fig_3}
\vspace{-5pt}
\end{figure*}

\noindent\textbf{Video frame samplers.}
We evaluated our adaptive sampler against uniform sampling using an equal number of sampled frames, with results shown in Table~\ref{tbl:sampler}. For videos without shot changes~\cite{xiao2021next, li2024mvbench}, both sampling methods perform similarly; for example, on NExT-QA, where only 1\% of the data contains shot changes, our adaptive sampling showed a modest improvement of 0.1\%. This comparable performance is expected, as frames within the same scene tend to be visually similar. However, the adaptive sampler demonstrates notable advantages on longer videos containing shot changes, achieving a 1\% improvement on MLVU~\cite{zhou2024mlvu}. Further analysis of MLVU task-specific performance reveals significant improvements in tasks that heavily rely on key frames, such as anomaly detection, plot understanding, and needle-in-the-haystack scenarios. These results validate both the effectiveness of our proposed AVS and its robustness, maintaining strong performance even in videos without shot changes. In Figure~\ref{fig:fig_3}, we demonstrate an example showing the different sampling results between uniform sampling and our AVS. As can be seen from the Fig~\ref{fig:fig_3}, uniform sampling picks many similar video frames and waste the token budget while our AVS can pick the key frames that are not overlapped with each other and successfully localize the frame showing a women holding a phone.

\noindent\textbf{Video on Compressor.}
We first conduct the ablation study on our SVC design. We sample 32 frames as input and utilize the proposed SVC with a default compression ratio of 64× (4×4×4 for temporal, width, and height dimensions) across all ablations unless otherwise specified. 

\noindent\textbf{Compressor Architecture.} 
We evaluated our proposed SVC method against two baseline approaches: average pooling and Perceiver-based down sampling~\cite{jaegle2021perceiver}, with results presented in Table~\ref{tbl:arch}. Our AE-based method demonstrates superior performance compared to both baselines at the same compression ratio (64×), validating the effectiveness of our SVC approach for video information compression. Notably, our method outperforms average pooling even with half number of tokens which suggests that simply increasing the number of tokens doesn't necessarily improve video representation. Rather, the key to effective video representation lies in eliminating redundant information while preserving essential spatial-temporal features. While the Perceiver approach achieves comparable performance to our method, it requires substantially more memory due to its cross-attention operations between queries and the numerous video tokens generated by the visual encoder.

\begin{table}[t]
\footnotesize
\centering
\begin{tabular}{lccccc}
\toprule
Method      & EgoS.     & MLVU          & MVbench       & NextQA      &  \\ 
\midrule
Random Init & 59.2          & 49.9          & 47.2          & 72.8        &  \\
Pre-train   & \textbf{63.6} & \textbf{51.9} & \textbf{50.0} & \textbf{77.0} & \\
\bottomrule
\end{tabular}
\caption{Experimental results of our method with pre-trained SVC and random initialized counterpart.}
\label{tbl:pretrain}
\end{table}

\begin{table}[t]
\footnotesize
\centering
\begin{tabular}{lcccc}
\toprule
Constrain & EgoS.         & MLVU          & MVBench       & NExT-QA       \\ 
\midrule
AE      & \multicolumn{4}{l}{Does not converge}                         \\
VAE     & \textbf{63.7} & 48.4          & 44.6          & 74.7          \\
APool+Res& 63.6 & \textbf{51.9} & \textbf{50.0} & \textbf{77.1} \\
\bottomrule
\end{tabular}
\caption{The effectiveness of different constraints in the autoencoder pretraining of SVC.}
\label{tbl:latent}
\end{table}

\noindent\textbf{Compression ratio.} 
In Table~\ref{tbl:cmp_ratio}, we analyze how different compression ratios affect benchmark performance. As expected, less aggressive compression yields better performance overall. However, when examining more aggressive compression scenarios, we discovered that spatial compression has a smaller negative impact on model performance compared to temporal compression. This difference might be attributed to the auto-encoder loss function's effectiveness in preserving spatial semantics during information compression. Furthermore, our experiments reveal that balanced spatial-temporal compression ratios perform better than the spatially or temporally heavy pooling approaches under the same total compression ratio. For instance, a balanced 4$\times$4$\times$4 compression strategy significantly outperforms the more asymmetric 1$\times$4$\times$4 configuration.

\noindent\textbf{AE pre-training.} We show that the autoencoder-based pre-training is necessary and critical for the compressor. As shown in Table~\ref{tbl:pretrain}, AE pre-training consistently leads to 2\%-4\% performance boosts on all benchmarks. it is worth mentioning that training the AE with long-form videos that has shot changes is critical.

\noindent\textbf{Latent space constraint.} We investigate the impact of different constraints on the latent space of autoencoders (AE) during pre-training, and results are shown in Table~\ref{tbl:latent}. Initially, we observed that when the compressor is pre-trained without any constraints, there are significant spikes in training loss, leading to feature alignment failure with LLMs. This result arises because the absence of constraints limits the generalization of the learnable compressor, potentially causing it to encode unseen data into gaps or "holes" in the encoder's feature space, resulting in a loss of meaningful representation. Introducing Gaussian distribution constraints on the latent space, as in VAE~\cite{kingma2013auto}, partially resolves the issue but still yields suboptimal results. One possible reason for this is the increased difficulty in learning feature representations due to the inherent nondeterminism of the features. In contrast, our proposed residual constraints demonstrate superior performance.

\section{Conclusion}
In this work, we present a novel advancement in video representation learning through our novel dual-component approach. The proposed adaptive video sampler effectively reduces spatial and temporal redundancies by leveraging contextual information, while our VAE-based spatiotemporal video compressor further optimizes video features without significant information loss. The experimental results across multiple video benchmarks demonstrate that our method not only achieves state-of-the-art performance, but also provides a more efficient solution for long-form video understanding. This work opens up new possibilities for processing and analyzing long video content in resource-constrained environments, marking an important step forward in the field of video understanding with large multimodal models. Future work could explore the adaptation of our method to real-time applications and investigate its potential in other multimedia domains.

{
    \small
    \bibliographystyle{ieeenat_fullname}
    \bibliography{main,supp}
}

\clearpage  
\appendix


\section{Implementation Details}
\subsection{MLLM Architecture}

We use the ViT-G/14~\cite{vit} pre-trained in the CLIP style~\cite{clip,openclip} as the visual encoder to extract visual token features from sampled frames. The extracted features are subsequently passed to the pre-trained compressor for feature compression. The compressed features are then mapped into the LLM input space using an 2-layers MLP projector~\cite{llava15}. Following the Llava-OneVision~\cite{llavaov}, we employ QwenV2-7B~\cite{qwen2} as our LLM.

\subsection{Adaptive Frame Sampler}
In this work, we employ TransNet V2~\cite{soucek2024transnet} as the shot boundary detector due to its superior performance and efficiency, enabling real-time processing. Frames are sampled at 10 FPS as input to TransNet V2~\cite{soucek2024transnet}. We set the window size for the non-maximum suppression algorithm to 3 seconds.

\begin{figure*}[t]
\centering
\includegraphics[width=\linewidth]{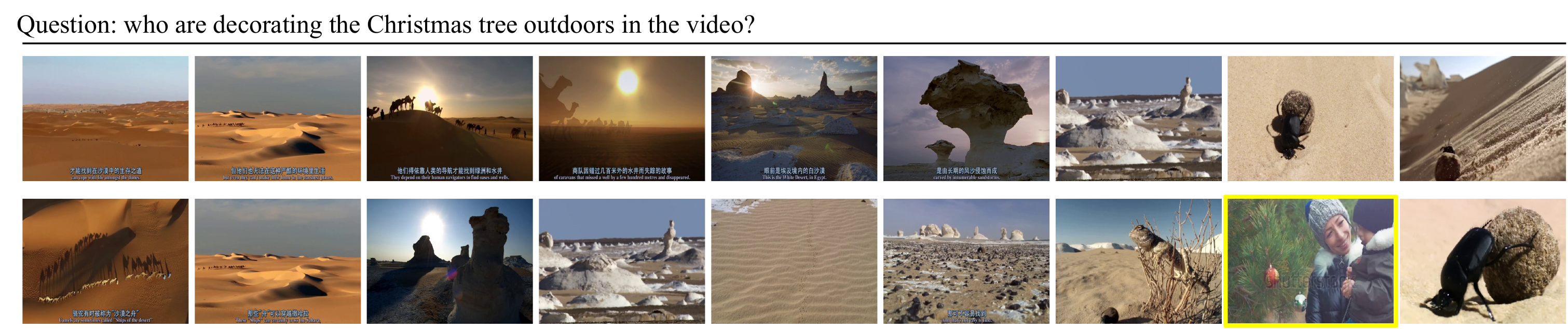}
  \caption{Examples of sampled frames using uniform sampling (top row) compared to our AVS (bottom row). AVS successfully locates the key frame to answer the question.}
\label{fig:fig_1}
\end{figure*}

\begin{figure*}[t]
\centering
\includegraphics[width=\linewidth]{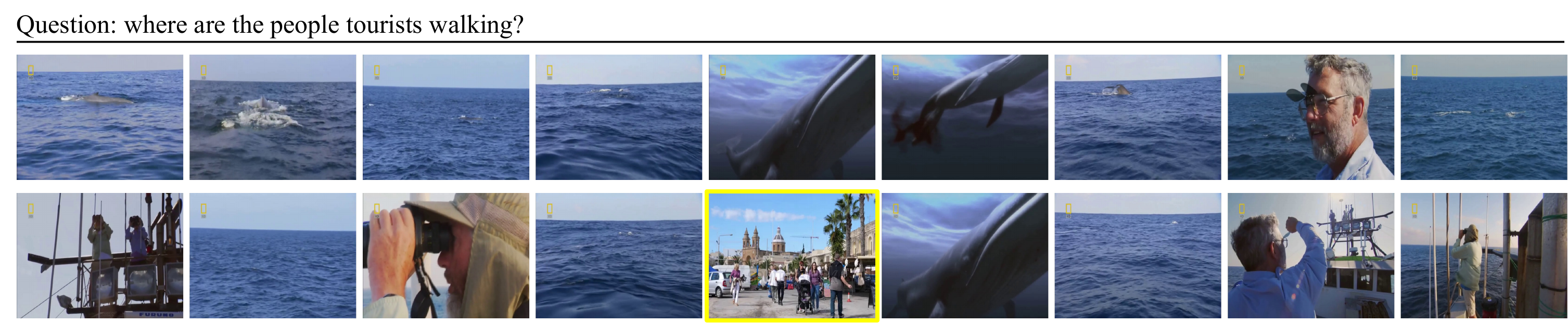}
  \caption{Examples of sampled frames using uniform sampling (top row) compared to our AVS (bottom row). AVS successfully locates the key frame to answer the question.}
\label{fig:fig_2}
\end{figure*}

\subsection{Compressor and Decoder Architecture}
The design of our compressor primarily follows the architecture of state-of-the-art (Variable-)autoencoders for images~\cite{vqgan} and videos~\cite{magvit2}. To reduce the parameter size of the compressor, we decompose 3D convolutions into 2D spatial and 1D temporal convolutions, and incorporate bottleneck connections between them.

To be specific, the compressor consists of cascaded \textit{residual blocks}, where each residual block comprises two \textit{convolutional blocks}. Each convolutional block contains a spatial 2D convolutional layer, followed by a 1D temporal convolution and a SiLU activation function. We avoid using 3D convolutions since they introduce a higher number of parameters. The 2D convolutional layer and the 1D temporal convolution are connected in a bottleneck fashion to reduce parameter size: the 2D convolutional layer outputs features with a dimensionality of $d/4$, while the temporal compression maps the features back to dimensionality $d$, where $d$ represents the dimensionality of the features extracted from the visual encoder. Downsampling is achieved by applying a convolutional stride of 2 in the first convolutional block. Our $4 \times 4 \times 4$ compressor configuration contains three such residual blocks, where the first two blocks perform downsampling, and the overall structure has approximately 65 million parameters.
For the decoder, it follows a symmetric architecture to the encoder, with upsampling operations follow those used in VQGAN~\cite{vqgan}.



\begin{table}[t]
\centering

\begin{tabular}{@{}ccc@{}}
\toprule
Training Statge          & Dataset Name   & Sample number \\ \midrule
Stage 1                  & ShutterStock   & 2,827,000     \\ \midrule
\multirow{7}{*}{stage 2} & ShutterStock   & 5,113,000     \\
                         & Ego4d          & 1,975,861     \\
                         & BreakFast      & 20,175        \\
                         & AVA            & 183,580       \\
                         & Vatex          & 252,200       \\
                         & SSV2           & 168,931       \\
                         & Kinetics       & 315,709       \\
                         & total          & 8,029,456     \\ \midrule
\multirow{6}{*}{SFT}     & ShutterStock   & 836,750       \\
                         & NextQA         & 71,655         \\
                         & CLEVRER        & 152,572        \\
                         & PerceptionTest & 7,392          \\
                         & Ego4d          & 11,018         \\
                         & total          & 1,079,387       \\ \bottomrule
\end{tabular}
\caption{The composition of training data across different training stages.}
\label{tbl:data}
\end{table}
\vspace{-5pt}

\subsection{Training Setting}

\noindent\textbf{Training Data Mixture:} The details of the training data used at different training stages are provided in Table~\ref{tbl:data}.
\noindent\textbf{MLLM training} 
We train our model on NVIDIA H100 GPUs with a batch size of 512. The training process is divided into three stages. In the stage-1, we train the MLP projector to align the compressed visual tokens with the LLM input space. We keep the LLM, visual encoder, and compressor frozen. The model is optimized by using the AdamW optimizer with a learning rate of $1 \times 10^{-3}$. In the stage-2, we train MLLM to learn new visual knowledge. The LLM and the projector are jointly trained with a AdamW optimizer~\cite{adamw} with a learning rate of $1 \times 10^{-5}$. In the SFT stage, we fine-tune the MLLM to solve various video understanding tasks and align with human preferences using our SFT data. We train the LLM and the projector with a AdamW optimizer~\cite{adamw} with a learning rate of $1 \times 10^{-5}$. \textit{We use 32 frames as input for both training and inference. For the ablation study, the model is trained only during stage 1 and the SFT stage to reduce computational costs. When comparing our method to state-of-the-art approaches (Table 1 in the main paper), the model is trained across stage 1, stage 2, and the SFT stage.} 

\noindent\textbf{Compressor pretraining} We pre-train our compressor on the Stage 2 training data for one epoch, using a batch size of 512. We use the AdamW optimizer~\cite{adamw} with a learning rate of $2 \times 10^{-5}$ to optimize the model. Note that the visual encoder is kept frozen.

\section{Additional Experiment Results}
\subsection{Comparison of Video Frame Samplers} In Figure~\ref{fig:fig_1} and Figure~\ref{fig:fig_2}, we additionally present the results of uniform sampling and our Adaptive Video Sampler. Consistent with our observations in the main paper, uniform sampling selects many redundant video frames, inefficiently utilizing the token budget. In contrast, our AVS effectively samples diverse key frames, accurately capturing the frame relevant to the questions ``Who is decorating the Christmas tree outdoors in the video?'' and ``Where are the people tourists walking?'' These results further demonstrate the effectiveness of our proposed AVS  approach.

\end{document}